\documentclass[runningheads]{llncs}
\usepackage{graphicx}
\DeclareGraphicsExtensions{.pdf,.png,.jpg,.eps,}
\usepackage{epstopdf}
\usepackage{amsmath,amssymb} % define this before the line numbering.

\usepackage{fixltx2e}
\usepackage{multirow}
\usepackage{capt-of}
\usepackage{color}
\usepackage[colorlinks=true]{hyperref}
\usepackage{mysymbols}

\begin{document}
\title{Multimodal Dual Attention Memory for \\ Video Story Question Answering.} 
% Replace with your title

\titlerunning{Multimodal Dual Attention Memory for Video Story Question Answering.}
% Replace with a meaningful short version of your title
%
\author{Kyung-Min Kim\inst{1}\thanks{Work carried out at Seoul National University and Surromind Robotics}\orcidID{0000-0003-2426-2198} \and
Seong-Ho Choi\inst{2}\orcidID{0000-0002-7553-6761} \and
Jin-Hwa Kim\inst{3}\thanks{Work carried out at Seoul National University}\orcidID{0000-0002-0423-0415} \and
Byoung-Tak Zhang\inst{2,4}\orcidID{0000-0001-9890-0389}}

%
%Please write out author names in full in the paper, i.e. full given and family names. 
%If any authors have names that can be parsed into FirstName LastName in multiple ways, please include the correct parsing, in a comment to the volume editors:
%\index{Lastnames, Firstnames}
%(Do not uncomment it, because you may introduce extra index items if you do that, we will use scripts for introducing index entries...)
\authorrunning{K.M. Kim, S.H. Choi, J.H. Kim, and B.T. Zhang}
% Replace with shorter version of the author list. If there are more authors than fits a line, please use A. Author et al.
%

\institute{Clova AI Research, NAVER Corp, Seongnam 13561, South Korea \and
Computer Science and Engineering, Seoul National University, Seoul 08826, South Korea \and
SK T-brain, Seoul 04539, South Korea \and
Surromind Robotics, Seoul 08826, South Korea\\
\email{\{kmkim,shchoi,jhkim,btzhang\}@bi.snu.ac.kr}}
\maketitle              % typeset the header of the contribution
\begin{abstract}
We propose a video story question-answering (QA) architecture, Multimodal Dual Attention Memory (MDAM). The key idea is to use a dual attention mechanism with late fusion. MDAM uses self-attention to learn the latent concepts in scene frames and captions. Given a question, MDAM uses the second attention over these latent concepts. Multimodal fusion is performed after the dual attention processes (late fusion). Using this processing pipeline, MDAM learns to infer a high-level vision-language joint representation from an abstraction of the full video content. We evaluate MDAM on PororoQA and MovieQA datasets which have large-scale QA annotations on cartoon videos and movies, respectively. For both datasets, MDAM achieves new state-of-the-art results with significant margins compared to the runner-up models. We confirm the best performance of the dual attention mechanism combined with late fusion by ablation studies. We also perform qualitative analysis by visualizing the inference mechanisms of MDAM.

\keywords{video story QA \and visual QA \and attention mechanism \and multimodal learning \and deep learning}
\end{abstract}

\section{Introduction}

Question-answering (QA) on a video story based on multimodal content input is an emerging topic in artificial intelligence. 
In recent years, multimodal deep learning studies have been successfully improving QA performance for still images~\cite{Agrawal,KimJH01,Fukui,VQA1} and video along with supporting content like subtitles, scripts, plot synopses, \etc~\cite{KimKM01,Na,TGIF-QA,VideoQA01}. 
Please note that video story QA is more challenging than image QA for the following two reasons.
 
First, video story QA involves multimodal content aligned on time-series. 
The model must learn the joint representations among at least two multimodal contents and given questions, and those joint representations must consider dynamic patterns over the time-series. 
Therefore, the use of multimodal fusion methods such as concatenation \cite{CoVQA,VQA2} or Multimodal Bilinear Pooling \cite{Fukui,Na,KimJH02} along with time axis might be prohibitively expensive and have the risk of over-fitting.

Second, video story QA requires to extract high-level meanings from the multimodal contents, \ie, scene frames and captions segmented based on the consistency of story. 
However, scene frames and captions in a video are redundant, highly-complex, and sometimes ambiguous information for the task, although humans can easily reason and infer based on the understanding of the video storyline in an abstract-level. 
It implies that humans can successfully extract the latent variables related to the multimodal content, which are used by the process of reasoning and inference. 
These latent variables are conditioned on a given question to give a correct answer. 
However, the previous work on video story QA has focused on the understanding of raw scene frames and captions without modeling on the latent variable \cite{KimKM01,Na,TGIF-QA,VideoQA01}. 

\begin{figure}[!tb]
\centering
\includegraphics[height=6.5cm]{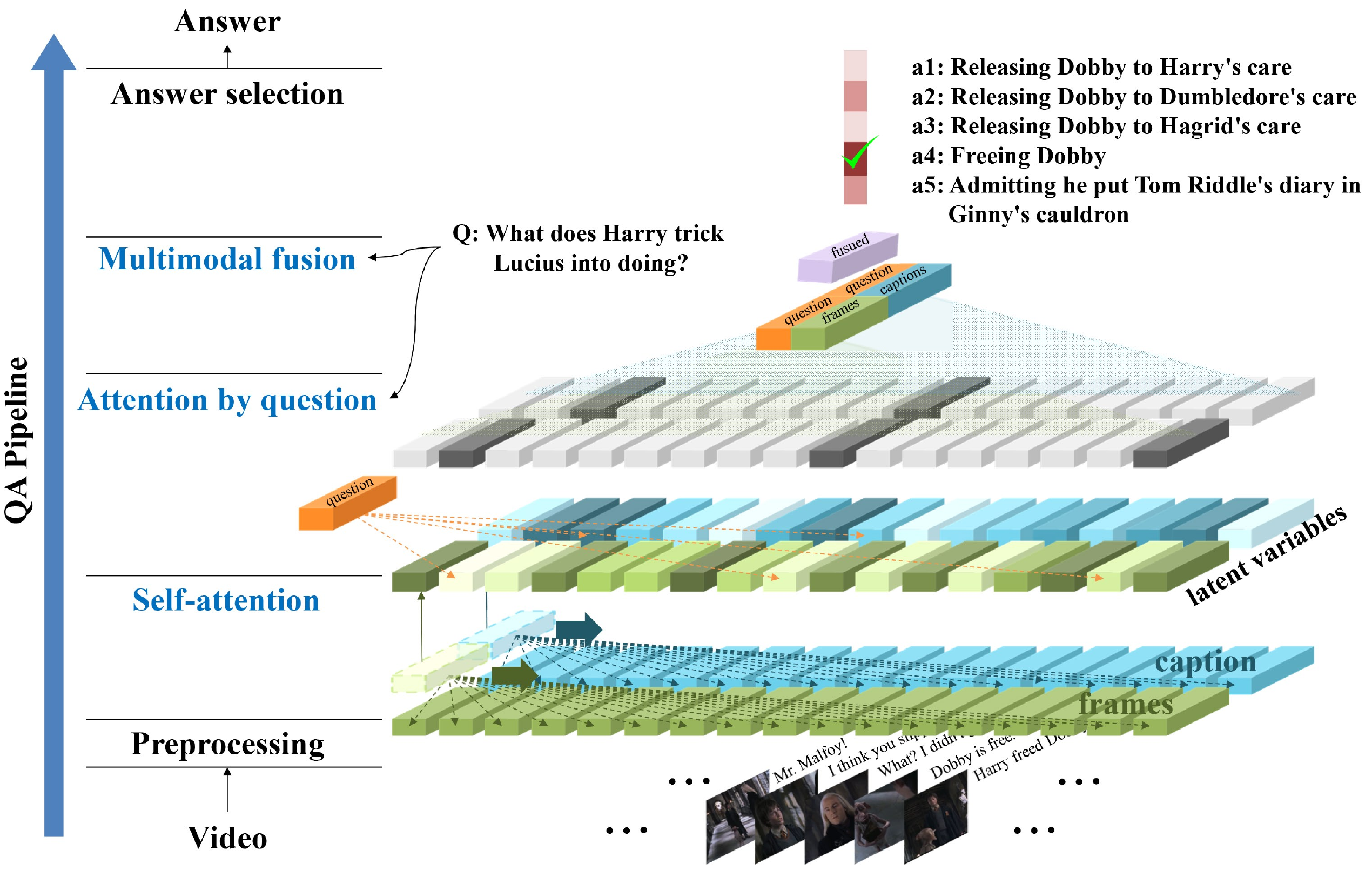}
\caption{The system architecture of Multimodal Dual Attention Memory (MDAM). 1) At the self-attention step, MDAM learns the latent variables of frames and captions based on the full video content. 2) For a given question, MDAM attend to the latent variables to remove unnecessary information. 3) At the multimodal fusion step, the question, caption, and frame information are fused using residual learning. During the whole inference process, multimodal fusion occurs only once.
}
\label{fig:mdam_abs}
\end{figure}

Here, we propose a novel model for video story QA task, Multimodal Dual Attention Memory (MDAM), which uses ResNet \cite{He}, GloVe \cite{GloVe}, positional encoding \cite{PE}, and casing features \cite{casing01} to represent scene frames and captions of a video. 
Then, using multi-head attention networks \cite{attn}, self-attention calculates the latent variables for the scene frames and captions. 
For a given question, the MDAM attends to the subset of the latent variables to compress scene frame and caption information to each single representation. 
After that, multimodal fusion occurs only once during the entire QA process, using the multimodal residual learning used in image QA \cite{KimJH01}. 
This learning pipeline consists of five submodules, preprocessing, self-attention, attention by question, multimodal fusion, and answer selection, which is learned end-to-end, supervised by given annotations. 
Fig. \hyperref[fig:mdam_abs]{1} shows the proposed model at an abstract level.

We evaluate our model on the large-scale video story QA datasets, MovieQA~\cite{MovieQA} and PororoQA~\cite{KimKM01}. 
The experimental results demonstrate two hypotheses of our model that 1) maximize QA related information through the dual attention process considering high-level video contents, and 2) multimodal fusion should be applied after high-level latent information is captured by our early process.

The main contributions of this paper are as follow: 
1) we propose a novel video story QA architecture with two hypotheses for video understanding; dual attention and late multimodal fusion, 
2) we achieve the state-of-the-art results on both PororoQA and MovieQA datasets, and our model is ranked at the first entry in the \textit{MovieQA Challenge} at the time of submission.

\section{Related Works}

\subsection{Video Story QA Datasets}

MovieQA aims to provide a movie dataset with high semantic diversity \cite{MovieQA}. 
The dataset consists of 408 movies and 14,944 multiple choices QAs. 
The dataset includes the stories of various genres such as action, fantasy, and drama; 
hence a QA model must be able to handle a variety of stories. 
The tasks of MovieQA can be divided into a text story QA mode (8,482 QA pairs) and a video story QA mode (6,462 QA pairs). 
The \textit{MovieQA Challenge} provides the evaluation server for test split so that participants can evaluate the performance of their models from this server.

Unlike MovieQA, PororoQA focuses on a coherent storyline \cite{KimKM01}. 
Since the videos are from a cartoon series, they provide more structured and simpler storylines. 
The dataset contains 27,328 scene descriptions and 8,834 multiple choices QA pairs with 171 videos of the children$'$s cartoon video series, \textit{‘Pororo’}.

\subsection{Video Story QA Models}
Deep Embedded Memory Networks (DEMN) \cite{KimKM01} replaces videos with generated text by combining scene descriptions and captions represented in a common linear embedding space. 
To solve QA tasks, DEMN evaluates all question-story sentence-answer triplets with the supervision of question and story sentence.

Read Write Memory Networks (RWMN) \cite{Na} fuse individual captions with the corresponding frames using Compact Bilinear Pooling \cite{Fukui} and store them in memory slots. 
Given the fact that each memory slot is not an independent entity, multi-layer convolutional neural networks are used to represent temporally adjacent slots. 
Our model provides a better solution to capture the latent variables of scene frames and captions through our dual attention mechanism for the full memory slots compared to the convolutional approach. 
Note that our multimodal fusion is applied to the latent variables instead of the early fusion in this work for high-level reasoning process.

ST-VQA applies attention mechanism on both spatial and temporal features of the videos \cite{TGIF-QA}. 
Unlike our proposed model, these attentions are only given to scene frames since the input of ST-VQA is short video clips such as GIFs without captions.
ST-VQA concatenates C3D \cite{C3D} and residual network features extracted from every interval of a video clip to obtain the spatial features. 
The model then calculates the temporal features of the intervals by feeding the spatial features into an LSTM. 
Given a question, attention mechanism is applied to both spatial and temporal features.

\begin{figure}[!tb]
\centering
\includegraphics[height=7.4cm]{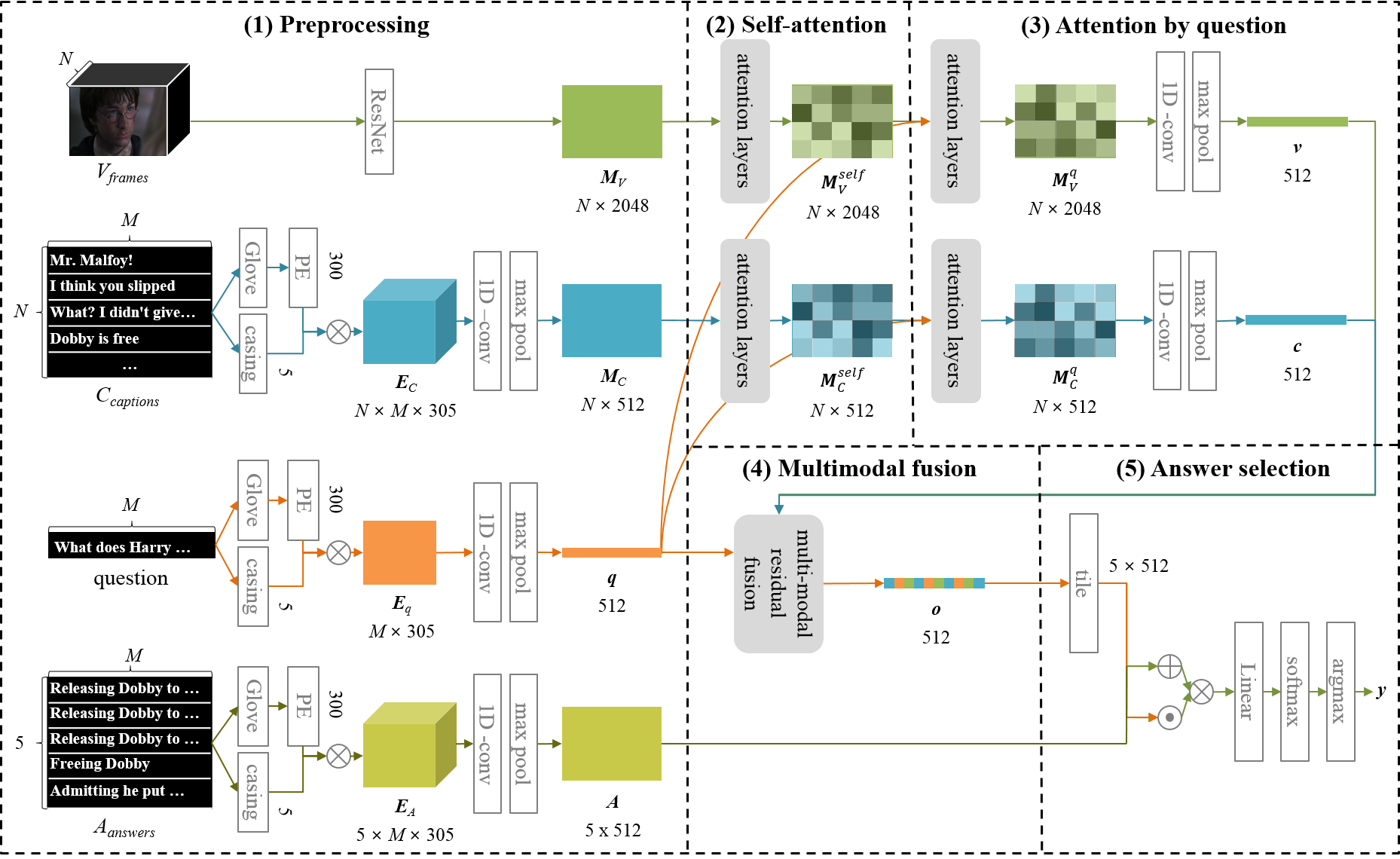}
\label{fig:MDAMfigure}
\caption{Five steps in our processing pipeline of Multimodal Dual Attention Memory for the video story QA task. (1) All given inputs are embedded as tensors and stored into long-term memory (Section \hyperref[sec:prep]{3.1}). (2) The frame tensor $\boldsymbol{M}_V^{self}$ and caption tensor $\boldsymbol{M}_C^{self}$ have latent variables of the frames and captions through the self-attention mechanism (Section \hyperref[sec:self]{3.2} and Fig. \hyperref[fig:attnlayers]{3}). (3) By using attention once again but with a question, the frames and captions are abstracted by the rank-1 tensors $\boldsymbol{v}$ and $\boldsymbol{c}$ (Section \hyperref[sec:attnQ]{3.3} and Fig. \hyperref[fig:attnlayers]{3}). (4) The fused representation $\boldsymbol{o}$ is calculated using residual learning fusion (Section \hyperref[sec:mmf]{3.4} and Fig. \hyperref[fig:mmfigure]{4}). (5) Finally, the correct answer sentence is selected with element-wise operations followed by the softmax classifier (Section \hyperref[sec:ans]{3.5}).} 
\end{figure}

\section{Multimodal Dual Attention Memory}\label{sec:MDAM}
Our goal is to build a video QA model that maximizes information needed for QA through attention mechanisms and fuses the multimodal information at a high-level of abstraction. We tackle this problem by introducing the two attention layers, which leverage the multi-head attention functions \cite{attn}, followed by the residual learning of multimodal fusion.

Fig. \hyperref[fig:MDAMfigure]{2} shows the overall architecture of our proposed model - Multimodal Dual Attention Memory(MDAM) for video story QA. 
The MDAM consists of five modules. 
1) The first module is the preprocessing module. All input including frames and captions of a given video is converted to the tensor formats.
2) In the self-attention module, the MDAM learns to obtain latent variables of the preprocessed frames and captions based on the whole video content. This process mimics a human who watches the full content of a video and then understands the story by recalling the frames and captions himself using the episodic buffer \cite{cogsci}.
3) In the attention by question module, the MDAM learns to give attention scores to find the relevant latent variables for a given question. It can be regarded as a cognitive process of finding points that contain answer information based on the understood story. 
4) These attentively refined frames and captions, and a question are fused using the residual function in the multimodal fusion module. 5) Finally, the answer selection module selects the correct answer by producing confidence score values over the five candidate answer sentences.

\subsection{Preprocessing}\label{sec:prep}
The input of the model is composed of 1) a sequence of frames \textit{V}\textsubscript{\textit{frames}} and 2) a sequence of captions \textit{C}\textsubscript{\textit{captions}} of a video clip \textit{I}\textsubscript{\textit{clip}} = \{\textit{V}\textsubscript{\textit{frames}}, \textit{C}\textsubscript{\textit{captions}}\}, 3) a question, and 4) a set of five candidate answer sentences \textit{A}\textsubscript{\textit{answers}} = (\textit{a}\textsubscript{1}, $\dotsc$, \textit{a}\textsubscript{5}). 
\textit{V}\textsubscript{\textit{frames}} and \textit{C}\textsubscript{\textit{captions}} consist of \textit{N} multiple frames and captions, \textit{V}\textsubscript{\textit{frames}} = (\textit{v}\textsubscript{1}, $\dotsc$, \textit{v}\textsubscript{\textit{N}}), \textit{C}\textsubscript{\textit{captions}} = (\textit{c}\textsubscript{1}, $\dotsc$, \textit{c}\textsubscript{\textit{N}}), where \textit{c}\textsubscript{\textit{i}} is a \textit{i}-th dialogue caption of the \textit{I}\textsubscript{\textit{clip}}, and \textit{v}\textsubscript{\textit{i}} is an image frame sampled at the midpoint between the start and end times of the caption \textit{c}\textsubscript{\textit{i}}. The value of the story length \textit{N} is fixed differently depending on the dataset used in this work. If the number of captions in the video is less than \textit{N}, zero padding is added. In Section \hyperref[sec:hyperparams]{4.2}, we will report the values of the hyperparameters. 

The main purpose of the preprocessing module is to transform the raw input as tensor formats, $\boldsymbol{M}_V$ $\in∈$$\mathbb{R}$\textsuperscript{\textit{N}$\times$2048}, $\boldsymbol{M}_C$ $\in∈$$\mathbb{R}$\textsuperscript{\textit{N}$\times$512}, $\boldsymbol{q}$ $\in∈$$\mathbb{R}$\textsuperscript{512}, $\boldsymbol{A}$ $\in∈$$\mathbb{R}$\textsuperscript{5$\times$512}, respectively, and store these in long-term memory, \eg, RAM. 
\subsubsection{Linguistic Inputs}
We first convert {\textit{C}\textsubscript{\textit{captions}}, question, \textit{A}\textsubscript{\textit{answers}} as word-level tensor representations, $\boldsymbol{E}_C$ $\in $ $\mathbb{R}$\textsuperscript{\textit{N}$\times$\textit{M}$\times$305}, $\boldsymbol{E}_q$ $\in $ $\mathbb{R}$\textsuperscript{\textit{M}$\times$305}, $\boldsymbol{E}_A$ $\in∈$$\mathbb{R}$\textsuperscript{5$\times$\textit{M}$\times$305}, respectively. \textit{M} is the fixed value denoting the maximum number of words in a sentence. Like the story length $N$, the value of $M$ depends on the dataset. For a sentence with less than \textit{M} words, zero padding is added. To represent each word of the inputs, we concatenate 300-D GloVe \cite{GloVe} with positional encoding \cite{PE}, and 5-D casing features.
 
\paragraph{GloVe and positional encoding.} Each word in the sentences is mapped to a GloVe embedding followed by positional encoding.
\begin{equation} \label{eq1}
e_i = g_i + p_i  \in \mathbb{R}^{300}
\end{equation}
where $g_i$ is GloVe embedding, and $p_i$ is the learnable embedding vector of the position index \textit{i}. $e_i$ is an output embedding.

\paragraph{Casing features.} As is used in the existing NLP studies \cite{casing01}, we add the following 5-D flag for each word representation. 1) A capitalization flag. This flag assigns the label True if at least one character of a word is upper-cased. 2) A numeric flag that assigns the label True if at least one character is numeric. 3) A personal pronouns flag that captures whether the word is a personal pronoun, \eg, she, he, they. 4) A unigram flag and 5) A bigram flag that indicate whether there is a unigram/bigram match between question and captions or question and candidate answer sentences. The casing feature is mapped to a five-dimensional zero-one vector.
\paragraph{}To obtain 512-D sentence-level tensor representations, we apply the shared 1-D convolution layers consisting of filters with varying window sizes $w^{e_1}_{conv}$ $\in∈$$\mathbb{R}$\textsuperscript{\textit{M}$\times$1$\times$1$\times$128}, $w^{e_2}_{conv}$ $\in∈$$\mathbb{R}$\textsuperscript{\textit{M}$\times$2$\times$1$\times$128}, $w^{e_3}_{conv}$ $\in∈$$\mathbb{R}$\textsuperscript{\textit{M}$\times$3$\times$1$\times$128}, $w^{e_4}_{conv}$ $\in∈$$\mathbb{R}$\textsuperscript{\textit{M}$\times$4$\times$1$\times$128} and max pooling operations to the word-level tensor representations, $\boldsymbol{E}_C$, $\boldsymbol{E}_q$, $\boldsymbol{E}_A$ \cite{KimY}.
\begin{gather}
\boldsymbol{M}_{C}[i] = \text{max}(\text{ReLU}(\text{conv}(\boldsymbol{E}_{C}[i,:,:], [w^{e_1}_{conv}, w^{e_2}_{conv}, w^{e_3}_{conv}, w^{e_4}_{conv}])))\\[1ex]
\boldsymbol{q} = \text{max}(\text{ReLU}(\text{conv}(\boldsymbol{E}_{q}, [w^{e_1}_{conv}, w^{e_2}_{conv}, w^{e_3}_{conv}, w^{e_4}_{conv}])))\\[1ex]
\boldsymbol{A}[j] = \text{max}(\text{ReLU}(\text{conv}(\boldsymbol{E}_{A}[j,:,:], [w^{e_1}_{conv}, w^{e_2}_{conv}, w^{e_3}_{conv}, w^{e_4}_{conv}])))
\end{gather}

\noindent where conv (input, filters) means the convolution layer, ReLU is the elementwise
ReLU activation \cite{ReLu}. Finally, the output tensors for the captions, question, answer sentences are $\boldsymbol{M}_C$ $\in∈$$\mathbb{R}$\textsuperscript{\textit{N}$\times$512}, $\boldsymbol{q}$ $\in∈$$\mathbb{R}$\textsuperscript{512}, $\boldsymbol{A}$ $\in∈$$\mathbb{R}$\textsuperscript{5$\times$512}, respectively, and they are stored into long-term memory.
\subsubsection{Visual Inputs}
The 2048-D sized activation output of 152-layer residual networks \cite{He} is used to represent {\textit{V}\textsubscript{\textit{frame}} as $\boldsymbol{M}_V$ $\in∈$$\mathbb{R}$\textsuperscript{\textit{N}$\times$2048}. It is stored in long-term memory.

\subsection{Self-attention}
\label{sec:self}
This module imports the frame tensor $\boldsymbol{M}_V$ $\in∈$$\mathbb{R}$\textsuperscript{\textit{N}$\times$2048}, and caption tensor $\boldsymbol{M}_C$ $\in∈$$\mathbb{R}$\textsuperscript{\textit{N}$\times$512} from the long-term memory as input. The output is the tensors $\boldsymbol{M}^{self}_V$ $\in∈$$\mathbb{R}$\textsuperscript{\textit{N}$\times$2048} and $\boldsymbol{M}^{self}_C$ $\in∈$$\mathbb{R}$\textsuperscript{\textit{N}$\times$512} that have latent values of the input by using attention layers \cite{attn}. The module provides separate attention to frames and captions.

Fig. \hyperref[fig:attnlayers]{3} (a) shows the process of the attention layers consisting of $L_{attn}$ identical layers \cite{attn}. Each layer has two sub-layers; 1) multi-head self-attention networks and 2) point-wise fully connected feed forward networks. There are a residual connection and layer normalization between each sub-layer. The $L_{attn}$ layers use different learning parameters for each layer.

\subsubsection{Multi-head Self-attention Networks}
In this sub-layer, each frame and caption can attend to all frames and captions
including itself to obtain a latent variable. It is achieved by selecting one pivot from the frames or captions and updating it using the attention mechanism.
Fig. \hyperref[fig:attnlayers]{3} (b) illustrates the detailed process. There are a pivot \textit{p} $\in \mathbb{R}^{d_k}$ and key set \textit{K} $\in \mathbb{R}^{N \times d_k}$. \textit{K} is the output of the previous layer or the input embedding, \ie, $\boldsymbol{M}_V$, $\boldsymbol{M}_C$, for the first layer. Each row vector of $K$ is a key whose latent variable is to be computed. $d_k$ is the dimension of the key, \ie, 512 or 2048. The pivot \textit{p} is selected from one of the $N$ keys of \textit{K}.

First, the networks project the pivot \textit{p} and \textit{N} keys to \textit{d}\textsubscript{\textit{proj}} dimensions
\textit{h} times, with different, learnable projection matrices. 
Then, for each projection, the weighted average using the scores obtained from the dot product-based attention by pivot $p$ aggregates $N$ keys.
\begin{gather}
head_i = \text{average}(\text{DotProdAttn}(pW^p_i, KW^K_i)))  
\in \mathbb{R}\textsuperscript{\textit{d}\textsubscript{\textit{proj}}} \label{eq:headi} \\[1ex] 
\text{where }\text{DotProdAttn}(x, Y) = \text{softmax}(xY^T/\sqrt{d_{proj}})Y \label{eq:dotprod} 
\end{gather}

\textit{h} outputs are concatenated and projected once again to become the updated key value $\tilde{K}[j,:]$ if the pivot \textit{p} is \textit{K}[\textit{j},:]. 
\begin{gather}
\tilde{K}[j,:] = (head\textsubscript{1} \otimes \cdots \otimes head\textsubscript{\textit{h}}) W_o \in \mathbb{R}^{d_k} \label{eq:concat}
\end{gather}

\noindent where $ \otimes$ denotes concatenation, and $W_o \in \mathbb{R}^{hd_{proj} \times d_k}$ is a projection matrix.
The networks change a pivot \textit{p} from {\textit{K}[\textit{1},:] to \textit{K}[\textit{N},:] and repeat the Eqn. (\hyperref[eq:headi]{5}) - (\hyperref[eq:concat]{7}) to obtain the updated key set $\tilde{K}$.

In this work, we use $h$=8, $d_{proj}$=64. In Section \hyperref[sec:quantitative]{4.3}, we will report the model performances according to the various $L_{attn}$ values.

\begin{figure} [!b]
\label{fig:attnlayers}
\centering
\includegraphics[height=4.4cm]{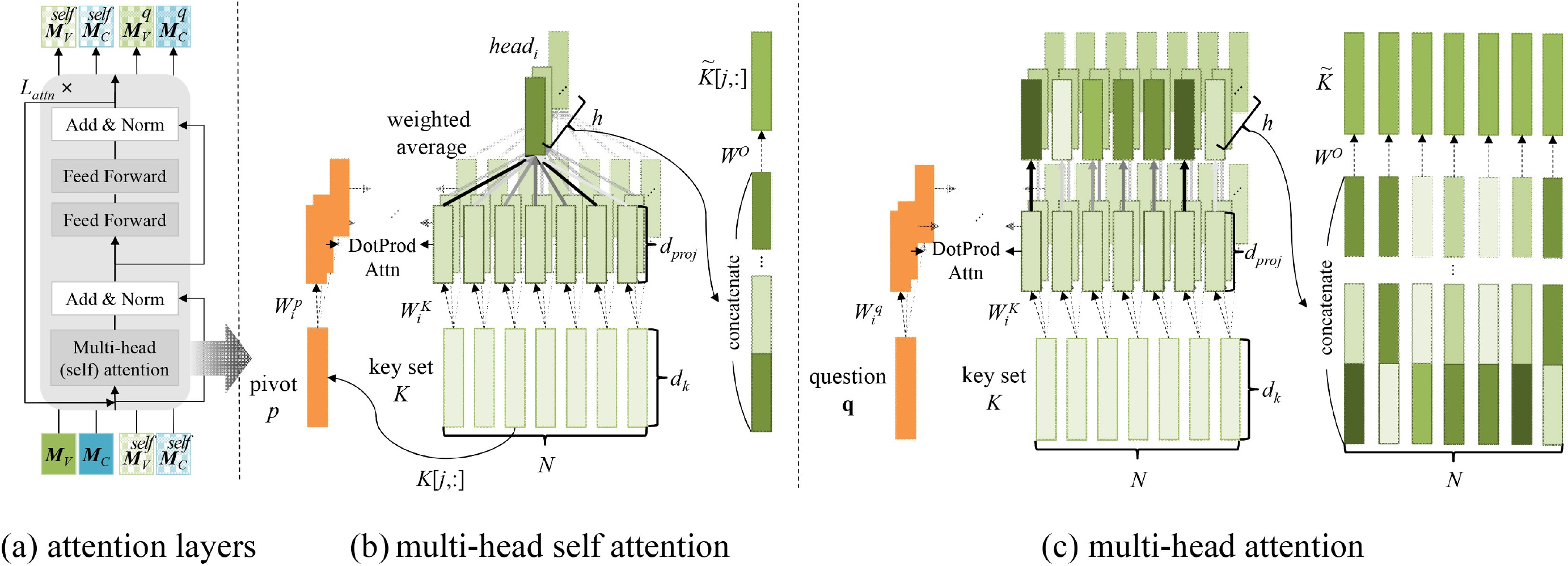}
\caption{(a) Illustration of the attention layers consisting of $L_{attn}$ identical layers. The self-attention module uses the multi-head self-attention networks while the attention by question module uses the multi-head attention networks. (b) The multi-head self-attention networks select a pivot \textit{p} from the key set \textit{K} to obtain the updated key set ${\tilde{K}}$. (c) The multi-head attention networks use the question \textbf{q} as a pivot to obtain the updated key set ${\tilde{K}}$.} 
\end{figure}

\subsubsection{Feed Forward Networks}
Fully-connected feed forward networks apply two linear transformations and a ReLU activation function separately and identically for every point of the input.
\begin{gather}
\text{FFN}(x) = \text{ReLU}(xW_1+b_1)W_2+b_2
\end{gather}
where \textit{x} is a point of the input. The dimension size of input and output is $d_k$, and the inner-layer has a dimension size of 2$d_k$.

\subsection{Attention by Question}
\label{sec:attnQ}
This module takes the final output tensors, $\boldsymbol{M}^{self}_V$, $\boldsymbol{M}^{self}_C$, of the self-attention module and again calculates the attention scores separately, by using the question. Attention information is aggregated using the 1-D convolutional neural networks to produce the output, $\boldsymbol{v}\in \mathbb{R}^{512}$ and $\boldsymbol{c}\in \mathbb{R}^{512}$, for the frames and captions, respectively. 
\subsubsection{Multi-head Attention Networks}
Like the self-attention module of Section \hyperref[sec:self]{3.2}, this module uses the attention layers consisting of $L_{attn}$ identical layers illustrated in Fig. \hyperref[fig:attnlayers]{3} (a) \cite{attn}. However, the attention layers differ from that of the self-attention module in that they have the multi-head attention networks inside. 

Fig. \hyperref[fig:attnlayers]{3} (c) shows the multi-head attention networks. 
The networks calculate the updated key set $\tilde{K}$ by applying attention to the key set $K$ as in Eqn. (\hyperref[eq:headi]{5}) - (\hyperref[eq:concat]{7}), but there are three differences when calculating Eqn. (\hyperref[eq:headi]{5}) and (\hyperref[eq:concat]{7}). 1) The networks use the question tensor $\boldsymbol{q}$ as a pivot by reading from the long-term memory. 2) The networks calculate the attention output values without average, \ie, \textit{head}\textsubscript{\textit{i}} =DotProdAttn(...) $\in \mathbb{R}\textsuperscript{\textit{N}$\times$\textit{d}\textsubscript{\textit{proj}}}$.
3) The output of the Eqn. (\hyperref[eq:concat]{7}) becomes $\tilde{K} \in \mathbb{R}^{N \times d_k}$ which is not a specific point of $\tilde{K}$.
\subsubsection{}
We denote the final output of the attention layers as $\boldsymbol{M}^q_V\in \mathbb{R}^{N \times 2048}$ and $\boldsymbol{M}^q_C\in \mathbb{R}^{N \times 512}$. Then, these are seperately aggregated using the 1-D convolutional neural networks and max pooling operation to get the outputs $\boldsymbol{v} \in \mathbb{R}^{512}$ and $\boldsymbol{c} \in \mathbb{R}^{512}$.
\begin{gather}
\mathbf{v} = \text{max}(\text{ReLU}(\text{conv}(\boldsymbol{M}^q_V, [w^{v_1}_{conv}, w^{v_2}_{conv}, w^{v_3}_{conv}, w^{v_4}_{conv}])))\label{eq:v}\\[1ex]
\mathbf{c} = \text{max}(\text{ReLU}(\text{conv}(\boldsymbol{M}^q_C, [w^{c_1}_{conv}, w^{c_2}_{conv}, w^{c_3}_{conv}, w^{c_4}_{conv}])))\label{eq:c}
\end{gather}
where $w^{v_i}_{conv} \in \mathbb{R}^{2048 \times i \times 1 \times 128}$ denote 1-D convolution filters of length \textit{i} for $\boldsymbol{M}^q_V$, and $w^{c_i}_{conv} \in \mathbb{R}^{512 \times i \times 1 \times 128}$ denote 1-D convolution filters of length \textit{i} for $\boldsymbol{M}^q_C$.

\subsection{Multimodal Fusion}
\label{sec:mmf}
During the entire QA process, multimodal fusion occurs only once in this module. The module fuses the refined frames $\boldsymbol{v} \in \mathbb{R}^{512}$, and captions $\boldsymbol{c} \in \mathbb{R}^{512}$, with the question $\boldsymbol{q} \in \mathbb{R}^{512}$ to output a single representation $\boldsymbol{o} \in \mathbb{R}^{512}$.  We borrow the idea of multimodal residual learning \cite{KimJH01}. 

\begin{figure} [!t]
\label{fig:mmfigure}
\centering
\includegraphics[height=4.8cm]{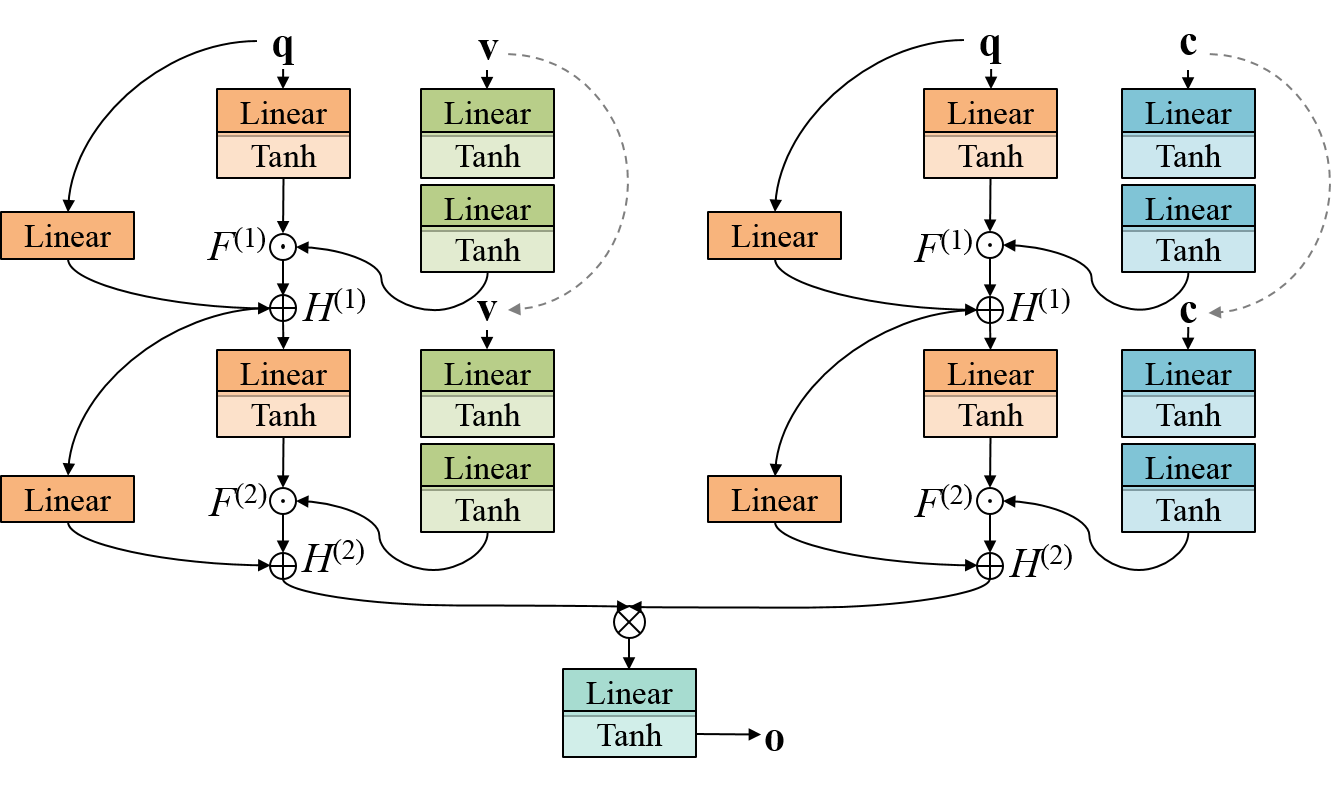}
\caption{A schematic diagram of the multimodal fusion module with the two deep residual blocks. The final output $\boldsymbol{o}$ is the concatenation of the outputs from the two residual blocks, $H^{(2)}$(question $\boldsymbol{q}$ - frame $\boldsymbol{v}$) and $H^{(2)}$(question $\boldsymbol{q}$ - caption $\boldsymbol{c}$), followed by a linear projection and \textit{tanh} activation.} 
\end{figure}

Fig. \hyperref[fig:mmfigure]{4} illustrates an example of our multimodal fusion module. The final output, $\boldsymbol{o}$, is the concatenation of the two deep residual blocks,  followed by linear projection and tanh activation. 
Each of the deep residual blocks consists of element-wise multiplication of question $\boldsymbol{q}$ and frames $\boldsymbol{v}$, or  $\boldsymbol{q}$ and captions $\boldsymbol{c}$, with residual connection.
\begin{gather}
\boldsymbol{o} = \sigma(W_o(H^{(L_m)}(\boldsymbol{q},\boldsymbol{v}) \otimes H^{(L_m)}(\boldsymbol{q},\boldsymbol{c}))) \\[1ex]
\text{where }H^{(L_m)}(\boldsymbol{q},x) = \boldsymbol{q}\prod_{l=1}^{L_m}W^{(l)}_{q}+ \sum\limits_{l=1}^{L_m} \{{F^{(l)}(H^{(l-1)},x)}\prod_{n=l+1}^{M}W^{(n)}_{q}\}\\[1ex]
F^{(l)}(H,x) = \sigma(HW^{(l)}_H) \odot \sigma(\sigma(xW^{(l)}_1)W^{(l)}_2 ) 
\end{gather}

\noindent where $L_m$ is the depth of the learning blocks. We use various values for $L_m$ in this work. $\otimes$ is the concatenation operation, $\sigma$ is the element-wise \textit{tanh} activation, and $\odot$ is the element-wise multiplication. $H^{(L_m)}(\boldsymbol{q},\boldsymbol{v})$ and $H^{(L_m)}(\boldsymbol{q},\boldsymbol{c})$ use different learning parameters.

\subsection{Answer Selection}
\label{sec:ans}
This module learns to select the correct answer sentence using the basic element-wise calculation between the output of multimodal fusion module, $\boldsymbol{o} \in \mathbb{R}^{512}$, and the answer sentence tensor, $\boldsymbol{A} \in \mathbb{R}^{5 \times 512}$, which is read from the long-term memory, follwed by the softmax classifier. 
\begin{gather}
O_A = (o_{tile} \odot \boldsymbol{A}) \otimes (o_{tile} \oplus \boldsymbol{A}) \\[1ex]
z = \text{softmax}(O_AW+b)
\end{gather}
Where $o_{tile} \in \mathbb{R}^{5 \times 512}$ is the tiled tensor of $\boldsymbol{o}$. $\oplus$ is the element-wise addition. $z \in \mathbb{R}^5$ is the confidence score vector over the five candidate answer sentences. Finally, we predict the answer $y$ with the highest score value, $y=argmax_{i\in[1,5]}(z_i)$.

\section{Experimental Results}

\subsection{Dataset}
The MovieQA dataset for the video QA mode consists of 140 movies with 6,462 QA pairs \cite{MovieQA}. Each question is coupled with a set of five possible answers; one correct and four incorrect answers. A QA model should choose a correct answer for a given question only using provided video clips and subtitles. The average length of a video clip is 202 seconds. If there are multiple video clips given in one question, we link them together into a single video clip. The number of QA pairs in train/val/test are 4318/886/1258, respectively. 

The PororoQA dataset has 171 episodes with 8,834 QA pairs \cite{KimKM01}. Like MovieQA, each question has one correct answer sentence and four incorrect answer sentences. One episode consists of a video clip of 431 seconds average length. For experiments, we split all 171 episodes into train(103 ep.)/val(34 ep.)/test(34 ep.) sets. The number of QA pairs in train/val/test are 5521/1955/1437, respectively. Unlike the MovieQA, the PororoQA has supporting fact labels that indicate which of the frames and captions of the video clip contain correct answer information, and description set. However, because our model does not use any supporting fact label or description, we do not use them in the experiment. 

\subsection{Experimental Setup}
\subsubsection{Pretrained Parameters} In the preprocessing module, \textit{ResNet-152} \cite{He} pre-trained with \textit{ImageNet} is used to encode the raw visual input, $\textit{V}\textsubscript{\textit{frame}}$. GloVe \cite{GloVe} pre-trained with \textit{Gigaword 5} and \textit{Wikipedia 2014} consisting of 6B tokens is used to encode the raw linguistic input, \textit{C}\textsubscript{\textit{captions}}, question, and \textit{A}\textsubscript{\textit{answers}}.
\subsubsection{Hyperparamters} \label{sec:hyperparams} For MovieQA, we limit the number of sentences per video clip to 40, \ie, $N$=40, and the number of words per sentence to 60, \ie, $M$=60. For PororoQA, we use $N$=20 and $M$=100. These are the maximum lengths of the sentences and words in each dataset.
Sentences or words below the given length are padded with zero values. We prevent the zero padding from participating in the error in the learning process.

The learnable parameters are initialized using the Xavier method \cite{Xavier} except for the pretrained models. The batch size is 16, and the number of epochs
is fixed to 160. Adam \cite{Adam} is used for optimization, and dropouts \cite{Dropout} are used for regularization. 

For learning rate and loss function, we empirically found that good parameters can be obtained by pre-training the model with the cross-entropy loss between the ground-truth one-hot vector $z_{gt}$ and prediction $z$ at a learning rate of 0.01 and then learning it again with the categorical hinge loss at a learning rate of 0.0001 from the best point. 

We train 20 different models and ensemble them using bayesian optimization.

\subsubsection{Baselines}
To compare the performance of each component, we conduct the ablation experiments with the following five model variants.
1) MDAM-MulFusion: model using element-wise multiplication instead of the residual learning function in the multimodal fusion module (self-attention is used).
2) MDAM-FrameOnly: model using only scene frames. 
3) MDAM-CaptOnly: model using only captions. 
4) MDAM-EarlyFusion: model that moves the position of the multimodal fusion module forward in the QA pipeline; thus the information flow goes through the following steps (i) preprocessing, (ii) multimodal fusion, (iii) self-attention, (iv) attention by question, (v) answer selection. 
The fusions of frames and captions occur \textit{N} times by fusing $\boldsymbol{M}_V$ and $\boldsymbol{M}_C$. 
5) MDAM-NoSelfAttn: model without the self-attention module. Furthermore, we measure the performance comparisons between our MDAM and other state-of-the-art models.

\begin{figure} [!b]
\label{fig:ablation}
\centering
\includegraphics[height=4.5cm]{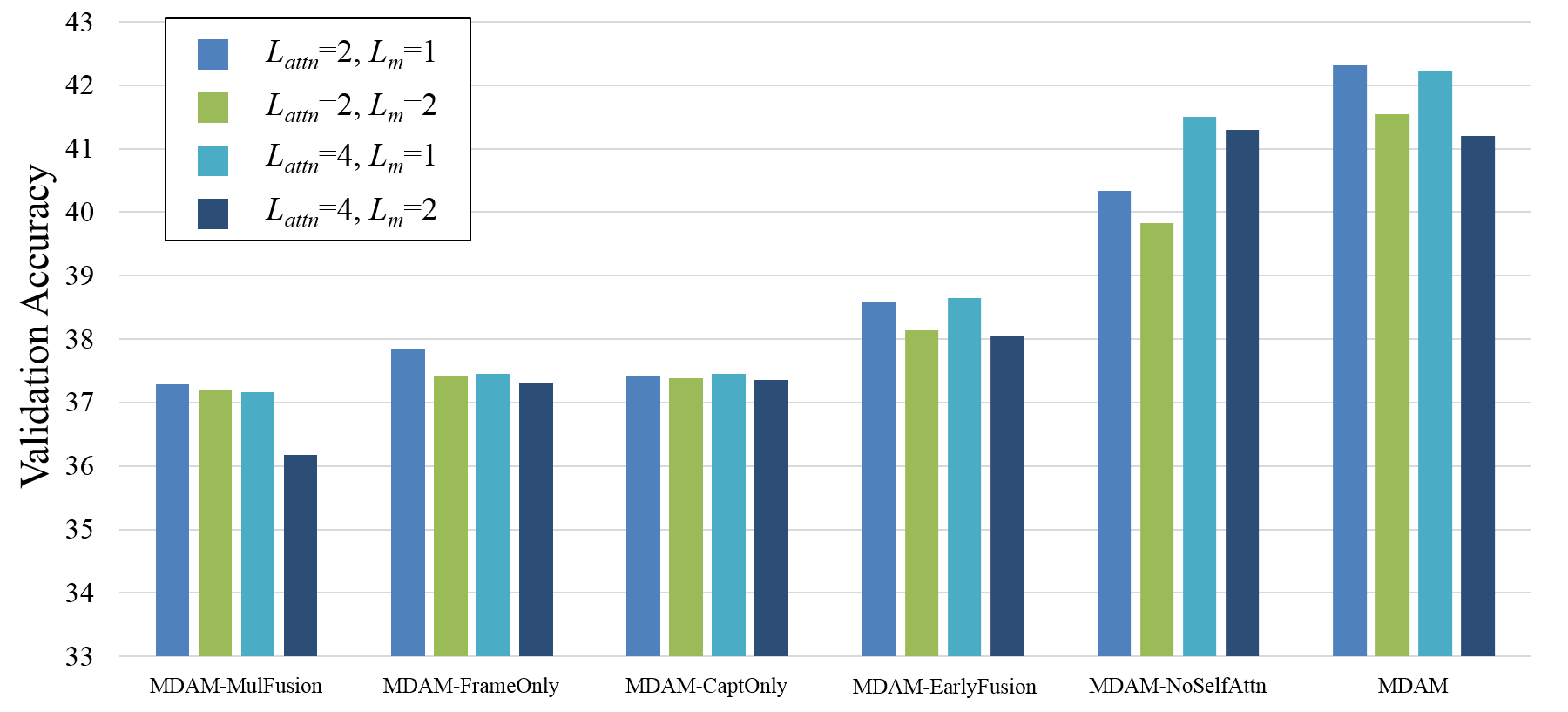}
\caption{The results of the model variants on the validation set of MovieQA. $L_{attn}$ denotes the number of layers in the attention networks. $L_m$ denotes the depth of the learning blocks in the multimodal fusion module.}
\end{figure}

\subsection{Quantitative Results} \label{sec:quantitative}
\subsubsection{MovieQA}
We report the experimental results of our model for validation and test sets.
We conduct the ablation experiments using the validation set to set the hyper-parameters of our models. Based on these results, we participated in the \textit{MovieQA Challenge}.
At the time of submission of the paper, our MDAM has recorded the highest accuracy of 41.41 \%
\paragraph{Ablation Experiments} Fig. \hyperref[fig:ablation]{5} shows the results of the ablation experiments.  
Due to the small size of the MovieQA data set, the overall performance pattern shows a tendency to decrease as the depth of the attention layers $L_{attn}$ and the depth of the learning blocks in the multimodal fusion module $L_m$ increase. 
Comparing the performance results by module, the models, in which multimodal fusions occur early in the QA pipeline (MDAM-EarlyFusion), shows little performance difference with the models, which use only sub-part of the video input (MDAM-FrameOnly, MDAM-CaptOnly). 
In addition, even if multimodal fusion occurs late, the performance is degraded where a simple element-wise multiplication is used as the fusion method (MDAM-MulFusion). 
Finally, the MDAM with self-attention through the full video content performs the best among our variant models.
These results imply the validity of our hypotheses of the model that 1) maximize the QA related information through the dual attention module and 2) fuse the multimodal information at a high-level of abstraction.
\paragraph{MovieQA Challenge} The \textit{MovieQA Challenge} provides a separate evaluation server for the test set so that participants can evaluate the performance of their models using the server. 
The evaluation is limited to once for every 72 hours.

Table \hyperref[fig:MovieQA]{1} shows the performance comparison with the other models released on the \textit{MovieQA Challenge} leaderboard. 
Our MDAM ($L_{attn}$=2, $L_m$=1) achieves 41.41\% and shows the performance gain of 2.38\% compared to the runner-up model, Layered Memory Network, which achieves 39.03\%.

\setlength{\tabcolsep}{4pt}
\begin{table}[!b]
\label{fig:MovieQA}
\begin{minipage}[c]{0.45\linewidth}
\caption{Performance comparison with other models proposed in the \textit{MovieQA Challenge} leaderboard of the video QA section.}
\label{table:challenge}
\begin{tabular}{lc}
\hline\noalign{\smallskip}
Method & test\\
\noalign{\smallskip}
\hline
\noalign{\smallskip}
LSTM+CNN  & 23.45\\
Simple MLP & 24.09\\
LSTM+discriminative CNN & 24.32\\
DEMN \cite{KimKM01} & 29.97\\
MuSM & 34.74 \\
RWMN \cite{Na} & 36.25\\
Local Avg. Pooling Networks & 38.16\\
Layered Memory Networks & 39.03\\
\hline
MDAM (ours) & {\bf 41.41}\\
\hline
\end{tabular}
\end{minipage}
\hspace{0.30cm}
\begin{minipage}[c]{0.50\linewidth}
\center
\label{table:PororoQA}
\caption{Performance comparison between other models proposed in \cite{KimKM01} and MDAM variants on the test set of PororoQA.}
\begin{tabular}{lc}
\hline\noalign{\smallskip}
Method & test\\
\noalign{\smallskip}
\hline
\noalign{\smallskip}
BoW V+Q & 34.2\\
W2V V+Q & 34.1\\
LSTM V+Q & 41.1\\
\hline
MDAM-MulFusion & {41.5}\\
MDAM-FrameOnly & {42.1}\\
MDAM-CaptOnly & {42.5}\\
MDAM-EarlyFusion & {46.1}\\
MDAM-NoSelfAttn & {47.3}\\
MDAM & {\bf 48.9}\\
\hline
\end{tabular}
\end{minipage}
\end{table}

\setlength{\tabcolsep}{1.4pt}

\begin{figure} [!ht]
\centering
\includegraphics[height=15.3cm]{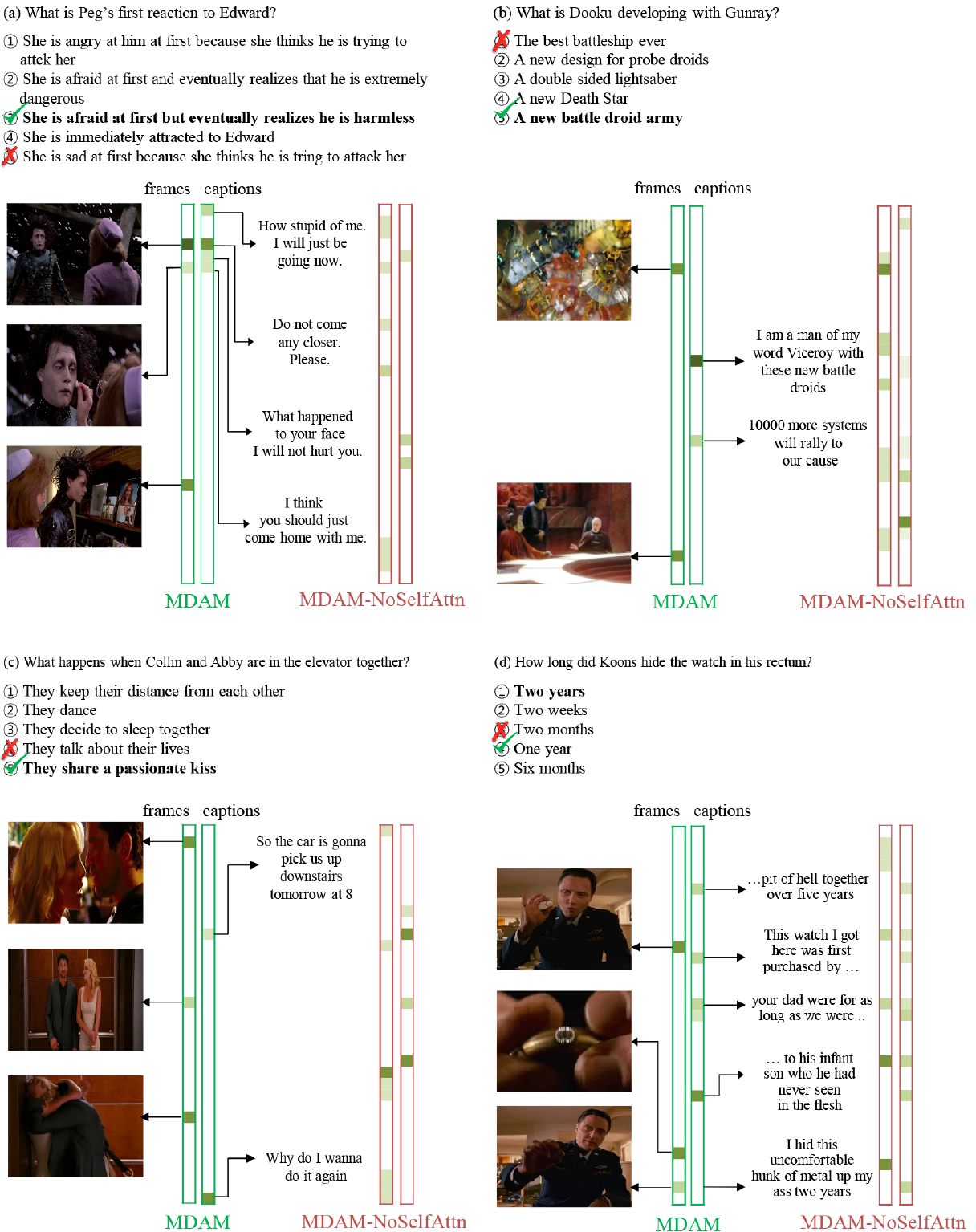}
\label{fig:qualitative}
\caption{Qualitative results for the MovieQA of MDAM with and without the self-attention module (MDAM and MDAM-NoSelfAttn, respectively). The successful cases are (a), (b), and (c), and the failure case is (d). Bold sentences are ground-truth answers. Green check symbols indicate the predictions of MDAM. Red cross symbols indicate the predictions of MDAM-NoSelfAttn. In each case, we show that which scene frames and captions are attended by the model for a given question.} 
\end{figure}

\subsubsection{PororoQA}
In Table \hyperref[table:PororoQA]{2}, we present the experimental results of our MDAM for the PororoQA dataset. 
The comparative models are the five MDAM variants and the existing baseline methods (BoW V+Q, W2V V+Q, LSTM V+Q) which do not use the descriptions and supporting fact labels like ours \cite{KimKM01}. 
As a result, MDAM achieves a state-of-the art performance (48.9 \%), marginally beating the existing methods.
Furthermore, we observe that the two hypotheses of MDAM are valid in PororoQA.
The self-attention module helps MDAM achieve better performance (48.9 \% for MDAM vs. 47.3 \% for MDAM-NoSelfAttn), and multimodal fusion with high-level latent information by our module performs better than early fusion baseline (46.1 \% for MDAM-EarlyFusion). All MDAM variants use $L_{attn}$=2 and $L_m$=1.

\subsection{Qualitative Results}
In this section, we visually analyze the inference mechanism of MDAM.
Fig. \hyperref[fig:qualitative]{6} shows the selected qualitative results of MDAM and MDAM without self-attention (MDAM-NoSelfAttn) for MovieQA.

Fig. \hyperref[fig:qualitative]{6} (a)-(c) show the successful examples of MDAM.
Given a question, MDAM solves the QA task correctly by attending to frames and captions containing answer-related information, which is performed by the attention by question module. 
Note that the model attends to frames and captions separately. 
It allows the model to focus on single modality one-by-one in the self-attention and attention by question modules for scene frames and captions in parallel. 

Fig. \hyperref[fig:qualitative]{6} (d) shows a challenging example of the MovieQA dataset.
The given video clip persists similar scenes with a long narrative by the character. 
These inputs make our MDAM to be challenging to select keyframes and corresponding captions which contain the information related to the given question, \ie time interval and the location of the watch.

\section{Concluding Remarks}
\label{sec:discuss}
We proposed a video story QA architecture, MDAM. 
The fundamental idea of ​​MDAM is to provide the dual attention structure that captures a high-level abstraction of the full video content by learning the latent variables of the video input, \ie, frames and captions, then, late multimodal fusion is applied to get a joint representation.   
We empirically demonstrated that our architectural choice is valid by showing the state-of-the-art performance on MovieQA and PororQA datasets.  
Exploring various alternative models in our ablation studies, we conjecture the following two points: 
1) The position of multimodal fusion in our QA pipeline is important to increase the performance.  
We learned that the early fusion models are easy to overfit, and the training loss fluctuates during a training phase due to many fusions occurred on time domain. 
On the other hand, the late fusion model were faster in convergence, leading to better performance results. 
2) For a given question, it is useful to attend to video content after self-attention.  
Because questions and scene frames are different modalities, \ie, language and vision, attending to a subset of the frames using a question tends to get a poor result if two hidden representations are not sufficiently aligned. 
Our self-attention module relieved this problem by calculating latent variables of frames and captions. 

\clearpage
\bibliographystyle{splncs04}
\bibliography{egbib}

\end{document}